\documentclass[runningheads]{llncs}

\usepackage{graphicx}
\usepackage{amsmath,amssymb} % define this before the line numbering.
\usepackage{makeidx}
\usepackage{graphicx}
\usepackage{amsmath,amssymb,amsfonts} % define this before the line numbering.

\usepackage{xspace}
\usepackage{subfig}

\def\eg{e.\,g.\xspace}
\def\ie{i.\,e.\xspace}
\def\etal{et~al.\xspace}
\def\wrt{w.\,r.\,t.\xspace}

\def\twod{\mbox{2-D}\xspace}
\def\threed{\mbox{3-D}\xspace}
\def\twothreed{\mbox{2-D/3-D}\xspace}
\def\xray{\mbox{X-ray}\xspace}

\newcommand{\volume}{V}
\newcommand{\fluoro}{I^\mathrm{FL}}

\newcommand{\pointV}{\mathbf{w}}

\newcommand{\pointI}{\mathbf{p}}

\newcommand{\motionVec}{\delta\mathbf{v}}
\newcommand{\transVec}{\delta{\boldsymbol{\nu}}}
\newcommand{\rotVec}{\delta{\boldsymbol{\omega}}}

\newcommand{\weightsCorrs}{\mathbf{s}}

\newcommand{\pointTarget}{\mathbf{q}}

\newcommand{\similarity}{\text{NGC}}

\newcommand{\featureVec}{\mathbf{f}}
\newcommand{\network}{\text{M}_{{\boldsymbol{\theta}}}}
\newcommand{\networkOpt}{\text{M}_{{\boldsymbol{\theta}}'}}
\newcommand{\networkParams}{{\boldsymbol{\theta}}}

\newcommand{\errorFunc}[1]{\text{PE}(#1)}
\newcommand{\regPPC}[1]{\text{RegPPC}(#1)}

\newcommand{\plane}{\mathrm{\Pi}}
\newcommand{\normal}{\mathbf{n}}
\newcommand{\projection}[2]{P_{#1}(#2)}

\newcommand{\gtReg}{\matr{T}^{\mathrm{GT}}}
\newcommand{\reg}{\matr{T}^\mathrm{est}}

\newcommand{\currReg}{\matr{T}}
\newcommand{\lastReg}{\hat{\matr{T}}}

\newcommand{\distance}[2]{\mathrm{d}(#1,#2)}
\newcommand{\vect}[1]{\mathbf{#1}}
\newcommand{\matr}[1]{\mathbf{#1}}
\newcommand{\transpose}{^\intercal}
\newcommand{\norm}[1]{\lVert #1 \rVert}

\newcommand{\set}[1]{\{#1\}}

\newcommand{\ppcl}{\mbox{PPC-L}\xspace}
\newcommand{\ppclp}{\mbox{PPC-L+}\xspace}
\newcommand{\ppcr}{\mbox{PPC-R}\xspace}
\newcommand{\ppcrm}{\mbox{PPC-RM}\xspace}
\newcommand{\ppcrmp}{\mbox{PPC-RM+}\xspace}

\newcommand{\realNumber}{\mathbb{R}}

\begin{document}
\title{Metric-Driven Learning of Correspondence Weighting for \twothreed Image Registration} 
% Replace with your title

\titlerunning{Learning of Correspondence Weighting for \twothreed Registration}
% Replace with a meaningful short version of your title
%
\author{
Roman~Schaffert\inst{1}
\and
Jian~Wang\inst{2}
\and
Peter~Fischer\inst{2}
\and
Anja~Borsdorf\inst{2}
\and
Andreas~Maier\inst{1,3}
}
%
%Please write out author names in full in the paper, i.e. full given and family names. 
%If any authors have names that can be parsed into FirstName LastName in multiple ways, please include the correct parsing, in a comment to the volume editors:
%\index{Lastnames, Firstnames}
%(Do not uncomment it, because you may introduce extra index items if you do that, we will use scripts for introducing index entries...)
\authorrunning{R. Schaffert et al.}
% Replace with shorter version of the author list. If there are more authors than fits a line, please use A. Author et al.
%

	\institute{Pattern Recognition Lab, Friedrich-Alexander Universität Erlangen-Nürnberg, Erlangen, Germany\\
	\email{roman.schrom.schaffert@fau.de}\\
	\and
	Siemens Healthineers AG, Forchheim, Germany\\
	\and
	Graduate School in Advanced Optical Technologies
	(SAOT), Erlangen, Germany
	}
\maketitle              % typeset the header of the contribution
\begin{abstract}

Registration of pre-operative \threed volumes to intra-operative \twod X-ray images is important in minimally invasive medical procedures. Rigid registration can be performed by estimating a global rigid motion that optimizes the alignment of local correspondences. However, inaccurate correspondences challenge the registration performance. To minimize their influence, we estimate optimal weights for correspondences using PointNet. We train the network directly with the criterion to minimize the registration error. We propose an objective function which includes point-to-plane correspondence-based motion estimation and projection error computation, thereby enabling the learning of a weighting strategy that optimally fits the underlying formulation of the registration task in an end-to-end fashion. For single-vertebra registration, we achieve an accuracy of 0.74$\pm$0.26 mm and highly improved robustness. The success rate is increased from 79.3\,\% to 94.3\,\% and the capture range from 3\,mm to 13\,mm.
\keywords{medical image registration \and 2-D/3-D registration \and deep learning \and \mbox {point-to-plane} correspondence model}
\end{abstract}
\section{Introduction}
\label{sec:intro}
Image fusion is frequently involved in modern \mbox{image-guided} medical interventions, typically augmenting \mbox{intra-operatively} acquired \twod \xray images with \mbox{pre-operative} \threed CT or MRI images. Accurate alignment between the fused images is essential for clinical applications and can be achieved using \twothreed rigid registration, which aims at finding the pose of a \threed volume in order to align its projections to \twod \xray images. Most commonly, \mbox{intensity-based} methods are employed~\cite{markelj2010review}, where a similarity measure between the \twod image and the projection of the \threed image is defined and optimized as \eg~described by Kubias~\etal~\cite{IMG08}. Despite decades of investigations, \twothreed registration remains challenging. The difference in dimensionality of the input images results in an \mbox{ill-posed} problem. In addition, content mismatch between the \mbox{pre-operative} and \mbox{intra-operative} images, poor image quality and a limited field of view challenge the robustness and accuracy of registration algorithms. Miao~\etal~\cite{DFM17} propose a \mbox{learning-based} registration method that is build upon the intensity-based approach. While they achieve a high robustness, registration accuracy remains challenging. 

The intuition of \twothreed rigid registration is to globally minimize the visual misalignment between \twod images and the projections of the \threed image.
Based on this intuition, Schmid and Ch{\^e}nes~\cite{segm2014Schmid} decompose the target structure to local shape patches and model image forces using Hooke's law of a spring from image block matching.
Wang~\etal~\cite{DRR17} propose a \mbox{point-to-plane} \mbox{correspondence (PPC)} model for \twothreed registration, which linearly constrains the global differential motion update using local correspondences. Registration is performed by iteratively establishing correspondences and performing the motion estimation.
During the intervention, devices and implants, as well as locally similar anatomies, can introduce outliers for local correspondence search (see Fig. \ref{fig:sample:td} and \ref{fig:sample:NGC}). Weighting of local correspondences, in order to emphasize the correct correspondences, directly influences the accuracy and robustness of the registration. 
An iterative reweighted scheme is suggested by Wang~\etal~\cite{DRR17} to enhance the robustness against outliers. However, this scheme only works when outliers are a minority of the measurements.

Recently, Qi~\etal~\cite{PND17} proposed the PointNet, a type of neural network directly processing point clouds. PointNet is capable of internally extracting global features of the cloud and relating them to local features of individual points. Thus, it is well suited for correspondence weighting in \twothreed registration. 
Yi~\etal~\cite{LFG18} propose to learn the selection of correct correspondences for wide-baseline stereo images. As a basis, candidates are established, \eg~using SIFT features. Ground truth labels are generated by exploiting the epipolar constraint. This way, an outlier label is generated. Additionally, a regression loss is introduced, which is based on the error in the estimation of a known essential matrix between two images. Both losses are combined during training. While including the regression
loss improves the results, the classification loss is shown to be important to find highly accurate correspondences.
The performance of iterative correspondence-based registration algorithms
(\eg~\cite{segm2014Schmid}, \cite{DRR17})
can be improved by learning a weighting strategy for the correspondences. 
However, automatic labeling of the correspondences is not practical for iterative methods as even correct correspondences may have large errors in the first few iterations. 
This means that labeling cannot be performed by applying a simple rule such as a threshold based on the ground truth position of a point.

In this paper, we propose a method to learn an optimal weighting strategy for the local correspondences for rigid \twothreed registration directly with the criterion to minimize the registration error, without the need of per-correspondence ground truth annotations.
We treat the correspondences as a point cloud with extended \mbox{per-point} features and use a modified PointNet architecture to learn global interdependencies of local correspondences according to the PPC registration metric.
We choose to use the PPC model as it was shown to enable a high registration accuracy as well as robustness~\cite{DRR17}. Furthermore, it is differentiable and therefore lends itself to the use in our training objective function.
To train the network, we propose a novel training objective function, which is composed of the motion estimation according to the PPC model and the registration error computation steps. It allows us to learn a correspondence weighting strategy by minimizing the registration error.
We demonstrate the effectiveness of the learned weighting strategy by evaluating our method on \mbox{single-vertebra} registration, where we show a highly improved robustness compared to the original PPC registration.

\section{Registration and Learned Correspondence Weighting}

In the following section, we begin with an overview of the registration method using the PPC model. 
Then, further details on motion estimation (see Sec.~\ref{sec:motionEstimation}) and registration error computation (see Sec.~\ref{sec:errorComputation}) are given, as these two steps play a crucial role in our objective function. 
The architecture of our network is discussed in Sec.~\ref{sec:architecture}, followed by the introduction of our objective function in Sec.~\ref{sec:objective}. 
At last, important details regarding the training procedure are given in Sec.~\ref{sec:training}.

\subsection{Registration Using Point-to-Plane Correspondences}
Wang~\etal~\cite{DRR17} measure the local misalignment between the projection of a \threed volume $\volume$ and the \twod fluoroscopic (live \xray) image $\fluoro$ and compute a motion which compensates for this misalignment.
Surface points are extracted from $\volume$ using the \threed Canny detector~\cite{CAE86}. 
A set of contour generator points~\cite{hartley03contGen} $\set{\pointV_i}$, \ie~surface points $\pointV_i\in\realNumber^3$ which correspond to contours in the projection of $\volume$, are projected onto the image as $\set{\pointI_i}$, \ie~a set of points $\pointI_i\in\realNumber^3$ on the image plane.
Additionally, gradient projection images of $\volume$ are generated and used to perform local patch matching to find correspondences for $\pointI_i$ in $\fluoro$. 
Assuming that the motion along contours is not detectable, the patch matching is only performed in the orthogonal direction to the contour. 
Therefore, the displacement of $\pointV_i$ along the contour is not known, as well as the displacement along the viewing direction. These unknown directions span the plane $\plane_i$ with the normal $\normal_i\in\realNumber^3$. After the registration, a point $\pointV_i$ should be located on the plane $\plane_i$.
To minimize the point-to-plane distances $\distance{\pointV_i}{\plane_i}$, a linear equation is defined for each correspondence under the small angle assumption. 
The resulting system of equations is solved for the differential motion $\motionVec\in\realNumber^6$, which contains both rotational components in the axis-angle representation $\rotVec\in\realNumber^3$ and translational components $\transVec\in\realNumber^3$, \ie~$\motionVec=(\rotVec\transpose, \transVec\transpose)\transpose$. 
The correspondence search and motion estimation steps are applied iteratively over multiple resolution levels. 
To increase the robustness of the motion estimation, the maximum correntropy criterion for regression (MCCR)~\cite{LMC15} is used to solve the system of linear equations~\cite{DRR17}.
The motion estimation is extended to coordinate systems related to the camera coordinates by a rigid transformation by Schaffert~\etal~\cite{MVD17}. 

The PPC model sets up a linear relationship between the local point-to-plane correspondences and the differential transformation, \ie a linear misalignment metric based on the found correspondences.
In this paper, we introduce a learning method for correspondence weighting, where the PPC metric is used during training to optimize the weighting strategy for the used correspondences with respect to the registration error.

\subsection{Weighted Motion Estimation}
\label{sec:motionEstimation}
Motion estimation according to the PPC model is performed by solving a linear system of equations  defined by $\matr{A}\in\realNumber^{N\times6}$ and $\vect{b}\in\realNumber^N$, where each equation corresponds to one point-to-plane correspondence and $N$ is the number of used correspondences. 
We perform the motion estimation in the camera coordinate system with the origin shifted to the centroid of $\set{\pointV_i}$. This allows us to use the regularized least-squares estimation
\begin{equation}
\motionVec = \underset{\motionVec'}{\arg\min}\left(\dfrac{1}{N}\norm{\matr{A}_s\motionVec'-\vect{b}_s}_2^2 + \lambda \norm{\motionVec'}_2^2\right)
\label{eq:LS}
\end{equation}
in order to improve the robustness of the estimation. 
Here, $\matr{A}_s=\matr{S}\cdot\matr{A}$, $\vect{b}_s=\matr{S}\cdot\vect{b}$ and $\lambda$ is the regularizer weight. The diagonal matrix $\matr{S}=\text{diag}(\vect{s})$ contains weights $\vect{s}\in\realNumber^N$ for all correspondences. As Eq.~\eqref{eq:LS} is differentiable \wrt $\motionVec'$, we obtain
\begin{equation}
\motionVec=\regPPC{\matr{A}, \vect{b},\weightsCorrs}=(\matr{A}_s\transpose\matr{A}_s+N\cdot\lambda \matr{I})^{-1}\matr{A}_s\transpose\vect{b}_s \enspace ,
\label{eq:LSClosedForm}
\end{equation}
where $\matr{I}\in\realNumber^{6\times6}$ is the identity matrix.
After each iteration, the registration $\currReg\in\realNumber^{4\times4}$ is updated as
\begin{equation}
\currReg =
\begin{pmatrix}
\cos(\alpha)\matr{I}+(1-\cos(\alpha)\vect{r}\vect{r}\transpose)+\sin(\alpha)[\vect{r}]_\times & \transVec \\ 0 & 1
\end{pmatrix}
\cdot \lastReg
\enspace ,
\label{eq:currReg}
\end{equation}
where $\alpha = \norm{\rotVec}$, $\vect{r} = \rotVec/\norm{\rotVec}$, $[\vect{r}]_\times\in\realNumber^{3\times3}$ is a skew matrix which expresses the cross product with $\vect{r}$ as a matrix multiplication and $\lastReg\in\realNumber^{4\times4}$ is the registration after the previous iteration~\cite{DRR17}.

\subsection{Registration Error Computation} 
\label{sec:errorComputation}
In the training phase, the registration error is measured and minimized via our training objective function. 
Different error metrics, such as the mean target registration error (mTRE) or the mean re-projection distance (mRPD) can be used. For more details on these metrics, see Sec.~\ref{sec:evalMetrics}.
In this work, we choose the projection error (PE)~\cite{GBD14}, as it directly corresponds to the visible misalignment in the images and therefore roughly correlates to the difficulty to find correspondences by patch matching for the next iteration of the registration method. The PE is computed as
\begin{equation}
e = \errorFunc{\currReg, \gtReg}=\dfrac{1}{M} \sum_{j=1}^M \norm{\projection{\currReg}{\pointTarget_j}-\projection{\gtReg}{\pointTarget_j}} \enspace ,
\label{eq:error}
\end{equation}
where a set of $M$ target points $\set{\pointTarget_j}$ is used and $j$ is the point index. $\projection{\currReg}{\cdot}$ is the projection onto the image plane under the currently estimated registration and $\projection{\gtReg}{\cdot}$ the projection under the \mbox{ground-truth} registration matrix $\gtReg\in\realNumber^{4\times4}$. Corners of the bounding box of the point set $\set{\pointV_i}$ are used as $\set{\pointTarget_j}$.

\begin{figure}[t]
\centering
  \includegraphics[width=1.0\textwidth]{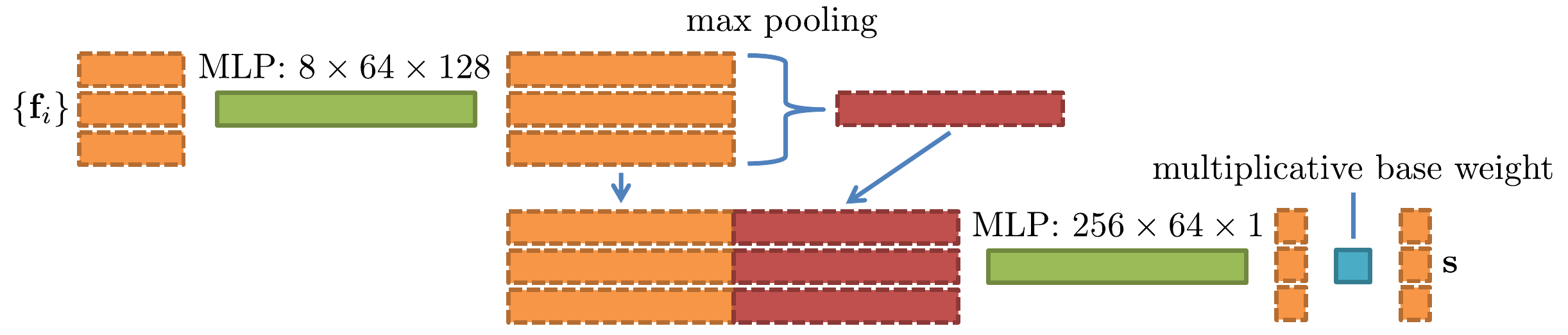}
     \caption{Modified PointNet~\cite{PND17} architecture used for correspondence weighting. Rectangles with dashed outlines indicate feature vectors (orange for local features, \ie containing information from single correspondences, and red for global features, \ie containing information from the entire set of correspondences). Sets of feature vectors (one feature vector per correspondence) are depicted as a column of feature vectors (three correspondences shown here). 
MLP denotes a multi-layer perceptron, which is applied to each feature vector individually.
     }
     \label{fig:architecture}
   \end{figure}

\subsection{Network Architecture} 
\label{sec:architecture}
We want to weight individual correspondences based on their geometrical properties as well as the image similarity, taking into account the global properties of the correspondence set.
For every correspondence, we define the features
\begin{equation}
\vect{f}_i =
\begin{pmatrix}
\pointV_i\transpose & \vect{n}_i\transpose & \distance{\pointV_i}{\plane_i} & \similarity_i& 
\end{pmatrix}\transpose \enspace ,
\label{eq:features}
\end{equation}
where $\similarity_i$ denotes the normalized gradient correlation for the correspondences, which is obtained in the patch matching step.

The goal is to learn the mapping from a set of feature vectors $\set{\vect{f}_i}$ representing all correspondences to the weight vector $\vect{s}$ containing weights for all correspondences, \ie~the mapping
\begin{equation}
\network: \set{\vect{f}_i} \mapsto \vect{s} \enspace , 
\end{equation}
where $\network$ is our network, and $\networkParams$ the network parameters.

To learn directly on correspondence sets, we use the PointNet~\cite{PND17} architecture and modify it to fit our task (see Fig.~\ref{fig:architecture}).
The basic idea behind PointNet is to process points individually and obtain global information by combining the points in a symmetric way, \ie~independent of order in which the points appear in the input~\cite{PND17}. 
In the simplest variant, the PointNet consists of a multi-layer perceptron (MLP) which is applied for each point, transforming the respective $\vect{f_i}$ into a \mbox{higher-dimensional} feature space and thereby obtaining a local point descriptor. 
To describe the global properties of the point set, the resulting local descriptors are combined by max pooling over all points, \ie~for each feature, the maximum activation over all points in the set is retained. 
To obtain per-point outputs, the resulting global descriptor is concatenated to the local descriptors of each point. 
The resulting descriptors, containing global as well as local information, are further processed for each point independently by a second MLP. 
For our network, we choose MLPs with the size of $8\times64\times128$ and $256\times64\times1$, which are smaller than in the original network~\cite{PND17}. 
We enforce the output to be in the range of $[0;1]$ by using a softsign activation function~\cite{elliott1993better} in the last layer of the second MLP and modify it to re-scale the output range from $(-1;1)$ to $(0;1)$.
Our modified softsign activation function $f(\cdot)$ is defined as
\begin{equation}
f(x) = \left(\dfrac{x}{1+|x|}+1\right)\cdot0.5 \enspace ,
\end{equation}
where $x$ is the state of the neuron.
Additionally, we introduce a global trainable weighting factor which is applied to all correspondences. 
This allows for an automatic adjustment of the strength of the regularization in the motion estimation step. 
Note that the network is able to process correspondence sets of variable size so that no fixed amount of correspondences is needed and all extracted correspondences can be utilized.

\subsection{Training Objective}
\label{sec:objective}
We now combine the motion estimation, PE computation and the modified PointNet to obtain the training objective function as
\begin{equation}
\boldsymbol{\theta}=\underset{\mathbf{\networkParams'}}{\arg\min}\dfrac{1}{K}\sum_{k=1}^K\errorFunc{\regPPC{\matr{A}_k, \vect{b}_k, \networkOpt(\set{\featureVec_i}_k)},\gtReg_k} \enspace ,
\label{eq:Objective}
\end{equation}
where $k$ is the training sample index and $K$ the overall number of samples. Equation \eqref{eq:LSClosedForm} is differentiable with respect to $\vect{s}$, Eq.~\eqref{eq:currReg} with respect to $\motionVec$ and Eq.~\eqref{eq:error} with respect to $\currReg$.
Therefore, gradient-based optimization can be performed on Eq.~\eqref{eq:Objective}.

Note that using Eq.~\eqref{eq:Objective}, we learn directly with the objective to minimize the registration error and no per-correspondence \mbox{ground-truth} weights are needed.
Instead, the PPC metric is used to implicitly assess the quality of the correspondences during the back-propagation step of the training and the weights are adjusted accordingly. In other words, the optimization of the weights is driven by the PPC metric.

\subsection{Training Procedure}
\label{sec:training}
To obtain training data, a set of volumes  $\set\volume$ is used, each with one or more \twod images $\set{\fluoro}$ and a known $\gtReg$ (see Sec.~\ref{sec:data}). For each pair of images, 60 random initial transformations with an uniformly distributed mTRE are generated~\cite{SEM05}. For details on the computation of the mTRE and start positions, see Sec.~\ref{sec:evalMetrics}.

Estimation of correspondences at training time is computationally expensive. 
Instead, the correspondence search is performed once and the precomputed correspondences are used during training. 
Training is performed for one iteration of the registration method and start positions with a small initial error are assumed to be representative for subsequent registration iterations at test time.
For training, the number of correspondences is fixed to 1024 to enable efficient batch-wise computations. The subset of used correspondences is selected randomly for every training step. Data augmentation is performed on the correspondence sets by applying translations, \mbox{in-plane} rotations and horizontal flipping, \ie reflection over the plane spanned by the vertical axis of the \twod image and the principal direction. For each resolution level, a separate model is trained.

\section{Experiments and Results}

\subsection{Data}
\label{sec:data}

\begin{figure}[t]
\centering
     \hfill
     \subfloat{%
       \includegraphics[width=0.32\textwidth]{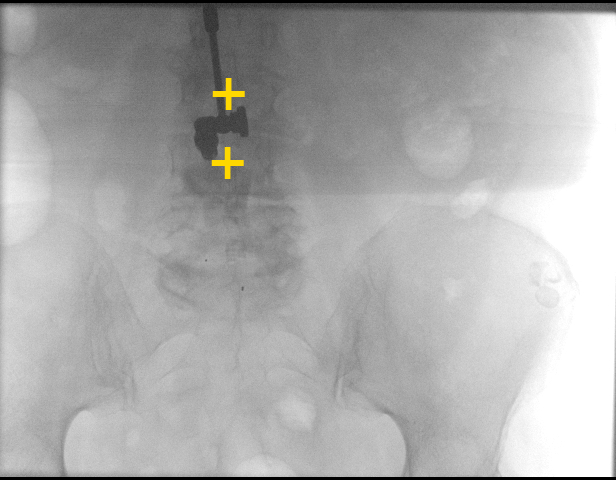}
     }
     \hfill
     \subfloat{%
       \includegraphics[width=0.32\textwidth]{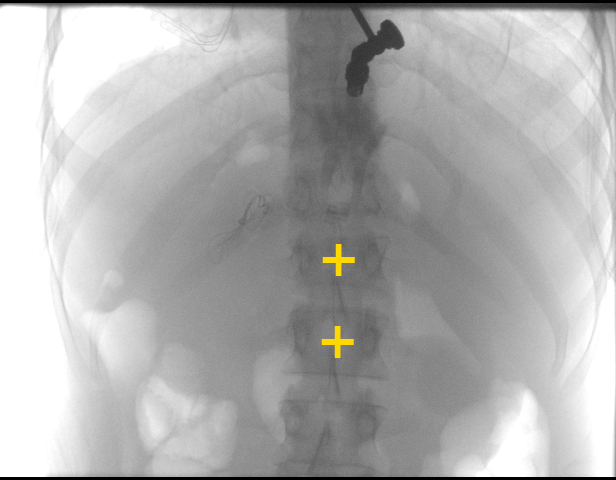}
     }
     \hfill
     \subfloat{%
       \includegraphics[width=0.32\textwidth]{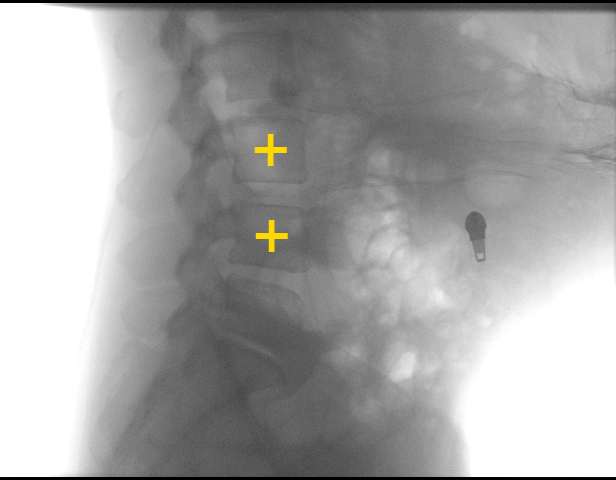}
     }
     \hfill
     \\
     \hfill
     \subfloat{%
       \includegraphics[width=0.32\textwidth]{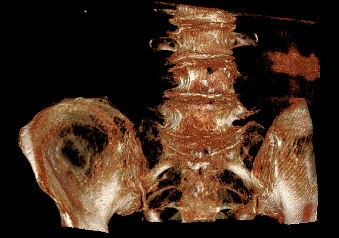}
     }
     \hfill
     \subfloat{%
       \includegraphics[width=0.32\textwidth]{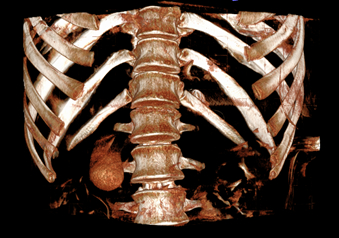}
     }
     \hfill
     \subfloat{%
       \includegraphics[width=0.32\textwidth]{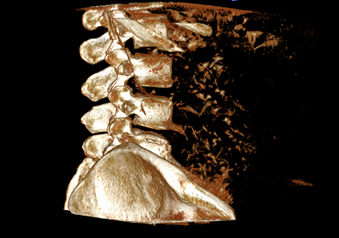}
     }
     \hfill
     \caption{Examples of \twod images used as $\fluoro$ (top row) and the corresponding \threed images used as $\volume$ (bottom row) in the registration evaluation. Evaluated vertebrae are marked by a yellow cross in the top row.}
     \label{fig:data}
   \end{figure}
  
We perform experiments for \mbox{single-view} registration of individual vertebrae. 
Note that \mbox{single-vertebra} registration is challenging due to the small size of the target structure and the presence of neighbor vertebrae. Therefore, achieving a high robustness is challenging.
We use clinical C-arm CT acquisitions from the thoracic and pelvic regions of the spine for training and evaluation. Each acquisition consists of a sequence of \twod images acquired with a rotating C-arm. These images are used to reconstruct the \threed volume. To enable reconstruction, the C-arm geometry has to be calibrated with a high accuracy (the accuracy is $\leq 0.16$\,mm for the projection error at the iso-center in our case). We register the acquired \twod images to the respective reconstructed volume and therefore the ground truth registration is known within the accuracy of the calibration.
Vertebra are defined by an axis-aligned volume of interest (VOI) containing the whole vertebra. Only surface points inside the VOI are used for registration. We register the projection images (resolution of $616\times480$ pixels, pixel size of 0.62\,mm) to the reconstructed volumes (containing around 390 slices with slice resolution of $512\times512$ voxels and voxel size of 0.49\,mm). 
To simulate realistic conditions, we add Poisson noise to all \twod images and rescale the intensities to better match fluoroscopic images.

The training set consists of  \mbox{19 acquisitions} with a total of  \mbox{77 vertebrae}.
For each vertebra,  \mbox{8 different} \twod images are used. An additional validation set of  \mbox{23 vertebrae} from  \mbox{6 acquisitions} is used to monitor the training process.
The registration is performed on a test set of 6 acquisitions. For each acquisition, \mbox{2 vertebrae} are evaluated and registration is performed independently for both the \mbox{anterior-posterior} and the lateral views.
Each set contains data from different patients, \ie~no patient appears in two different sets. The sets were defined so that all sets are representative to the overall quality of the available images, \ie~contain both pelvic and thoracic vertebrae, as well as images with more or less clearly visible vertebrae.
Examples of images used in the test set are shown in Fig.~\ref{fig:data}.

\subsection{Compared Methods}
\label{sec:comparedMethods}
We evaluate the performance of the registration using the PPC model in combination with the learned correspondence weighting strategy (PPC-L), which was trained using our proposed metric-driven learning method.
To show the effectiveness of the correspondence weighting, we compare PPC-L to the original PPC method. The compared methods differ in the computation of the correspondence weights $\vect{s}$ and the regularizer weight $\lambda$. For \ppcl, the correspondence weights $\vect{s}^\mathrm{L} = \network({\set{\featureVec})}$ and $\lambda = 0.01$ are used. For PPC, we set $\lambda = 0$ and the used correspondence weights $\vect{s}^\mathrm{PPC}$ are the $\similarity_i$ values of the found correspondences, where any value below $0.1$ is set to $0$, \ie~the correspondence is rejected. Additionally, the MCCR is used in the PPC method only. The minimum resolution level has a scaling of 0.25 and the highest a scaling of 1.0. For the PPC method, registration is performed on the lowest resolution level without allowing motion in depth first, as this showed to increases the robustness of the method. To differentiate between the effect of the correspondence weighting and the regularized motion estimation, we also consider registration using regularized motion estimation. We use a variant where the global weighting factor, which is applied to all points, is matched to the regularizer weight automatically by using our objective function (\ppcr). For the different resolution levels, we obtained a data weight in the range of $[2.0 ; 2.1]$. Therefore, we use $\lambda = 0.01$ and $\vect{s}^\mathrm{R} = 2.0 \cdot \vect{s}^\mathrm{PPC}$. Additionally, we empirically set the correspondence weight to $\vect{s}^\mathrm{RM} = 0.25 \cdot \vect{s}^\mathrm{PPC}$, which increases the robustness of the registration while still allowing for a reasonable amount of motion (\ppcrm).

\subsection{Evaluation Metrics}
\label{sec:evalMetrics}
To evaluate the registration, we follow the standardized evaluation methodology~\cite{SEM05,ROC13}. 
The following metrics are defined by van de Kraats~\etal~\cite{SEM05}:
\begin{itemize}
\item{\it Mean Target Registration Error:}
The mTRE is defined as the mean distance of target points under $\gtReg$ and the estimated registration $\reg\in\realNumber^{4\times4}$.
\item{\it Mean Re-Projection Distance (mRPD):}
The mRPD is defined as the mean distance of target points under $\gtReg$ and the \mbox{re-projection} rays of the points as projected under $\reg$.
\item{\it Success Rate (SR):}
The SR is the number of registrations with with a registration error below a given threshold.  As we are concerned with \mbox{single-view} registration, we define the success criterion as a mRPD $\leq$ 2\,mm.
\item{\it Capture Range (CR):}
The CR is defined as the maximum initial mTRE for which at least 95\% of registrations are successful.
\end{itemize}
Additionally, we compute the gross success rate (GSR)~\cite{DFM17} as well as a gross capture range (GCR) with a success criterion of a mRPD $\leq$ 10\,mm in order to further assess the robustness of the methods in case of a low accuracy.
We define target points as uniformly distributed points inside the VOI of the registered vertebra.
For the evaluation, we generate 600 random start transformations for each vertebra in a range \mbox{of 0\,mm - 30\,mm} initial mTRE using the methodology described by van de Kraats~\etal~\cite{SEM05}. 
We evaluate the accuracy using the mRPD and the robustness using the SR, CR GSR and GCR. 
 
\subsection{Results and Discussion}

\subsubsection{Accuracy and Robustness}
\begin{figure}[t]
\centering
     \subfloat[\label{fig:sample:td}]{%
       \includegraphics[width=0.23\textwidth]{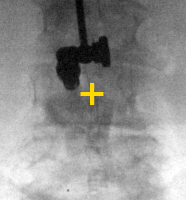}
     }
     \hfill
     \subfloat[\label{fig:sample:NGC}]{%
       \includegraphics[width=0.23\textwidth]{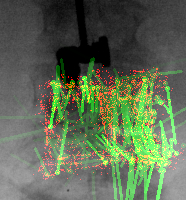}
     }
     \hfill
     \subfloat[\label{fig:sample:W}]{%
       \includegraphics[width=0.23\textwidth]{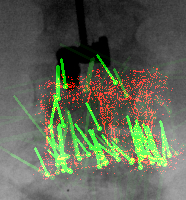}
     }
     \hfill
     \subfloat[\label{fig:sample:res}]{%
       \includegraphics[width=0.23\textwidth]{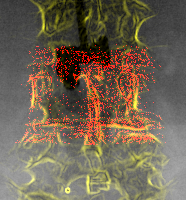}
     }
     \caption{Registration example: (a) shows $\fluoro$ with one marked vertebra to register. Red dots depict initially extracted (b,\,c) and final aligned (d) contour points. Green lines depict the same randomly selected subset of correspondences, whose intensities are determined by $\similarity_i$ (b) and learned weights (c). Final \ppcl registration result overlaid in yellow (d). Also see video in the supplementary material.
     }
     \label{fig:sample}
   \end{figure}

The evaluation results for the compared methods are summarized in Tab. \ref{tab:resBase}. We observe that \ppcl achieves the best SR of 94.3\,\% and CR of 13\,mm. Compared to PPC (SR of 79.3\,\% and CR of 3\,mm), \ppcr also achieves a higher SR of 88.1\,\% and CR of 6\,mm. For the regularized motion estimation, the accuracy decreases for increasing regularizer influence (0.79$\pm${0.22}\,mm for \ppcr and 1.18$\pm${0.42}\,mm for \ppcrm), compared to PPC (0.75$\pm$0.21\,mm) and \ppcl (0.74$\pm$0.26\,mm). A sample registration result using \ppcl is shown in Fig.~\ref{fig:sample:res}.
\begin{table}[b]
\centering
\caption{Evaluation results for the compared methods. The mRPD is computed for the 2\,mm success criterion and is shown as mean\,$\pm$\,standard deviation.}
\label{tab:angioRes}
\begin{tabular}{l|c|c|c|c|c}
\hline
Method    & mRPD    {[}mm{]} & SR {[}\%{]} & CR {[}mm{]} & GSR {[}\%{]} & GCR {[}mm{]}\\ 
\hline
PPC				& 0.75$\pm$0.21				& 79.3 & 3			& 81.8		& 3	 	\\
\ppcr			& {0.79}$\pm${0.22}			& 88.1		& 6		 	& 90.72	& 6 	\\
\ppcrm			& {1.18}$\pm${0.42}			& 59.6		& 4	 		& 95.1		& 20	\\
\bf{\ppcl}		& {{0.74}}$\pm$0.26	 		& 94.3		& 13	 	& 96.3		& 22	\\
\hline
\end{tabular}
\label{tab:resBase}
\end{table}

For strongly regularized motion estimation, we observe a large difference between the GSR and the SR. While for \ppcr, the difference is relatively small \mbox{(88.1\% vs. 90.7\%)}, it is very high for \ppcrm. Here a GSR of 95.1\,\% is achieved, while the SR is 59.6\,\%. This indicates that while the method is robust, the accuracy is low. Compared to the CR, the GCR is increased for \ppcl (22\,mm vs. 13\,mm) and especially for \ppcrm (20\,mm vs. 4\,mm). 
Overall, this shows that while some inaccurate registrations are present in \ppcl, they are very common for \ppcrm.

\subsubsection{Single Iteration Evaluation}
\begin{figure}[b]
\centering
     \subfloat[PPC]{%
       \includegraphics[width=0.32\textwidth]{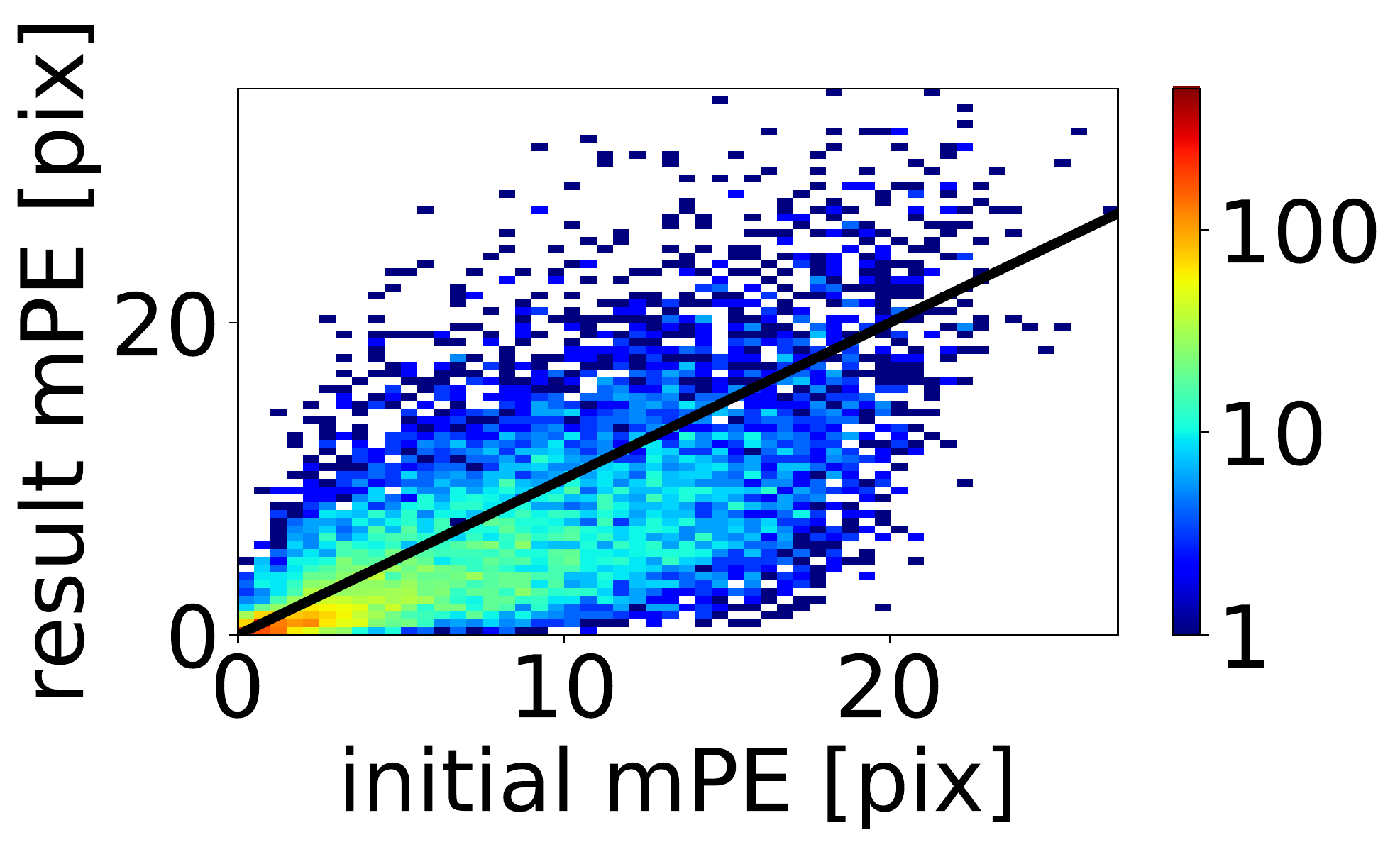}
     }
     \hfill
     \subfloat[\ppcr]{%
       \includegraphics[width=0.32\textwidth]{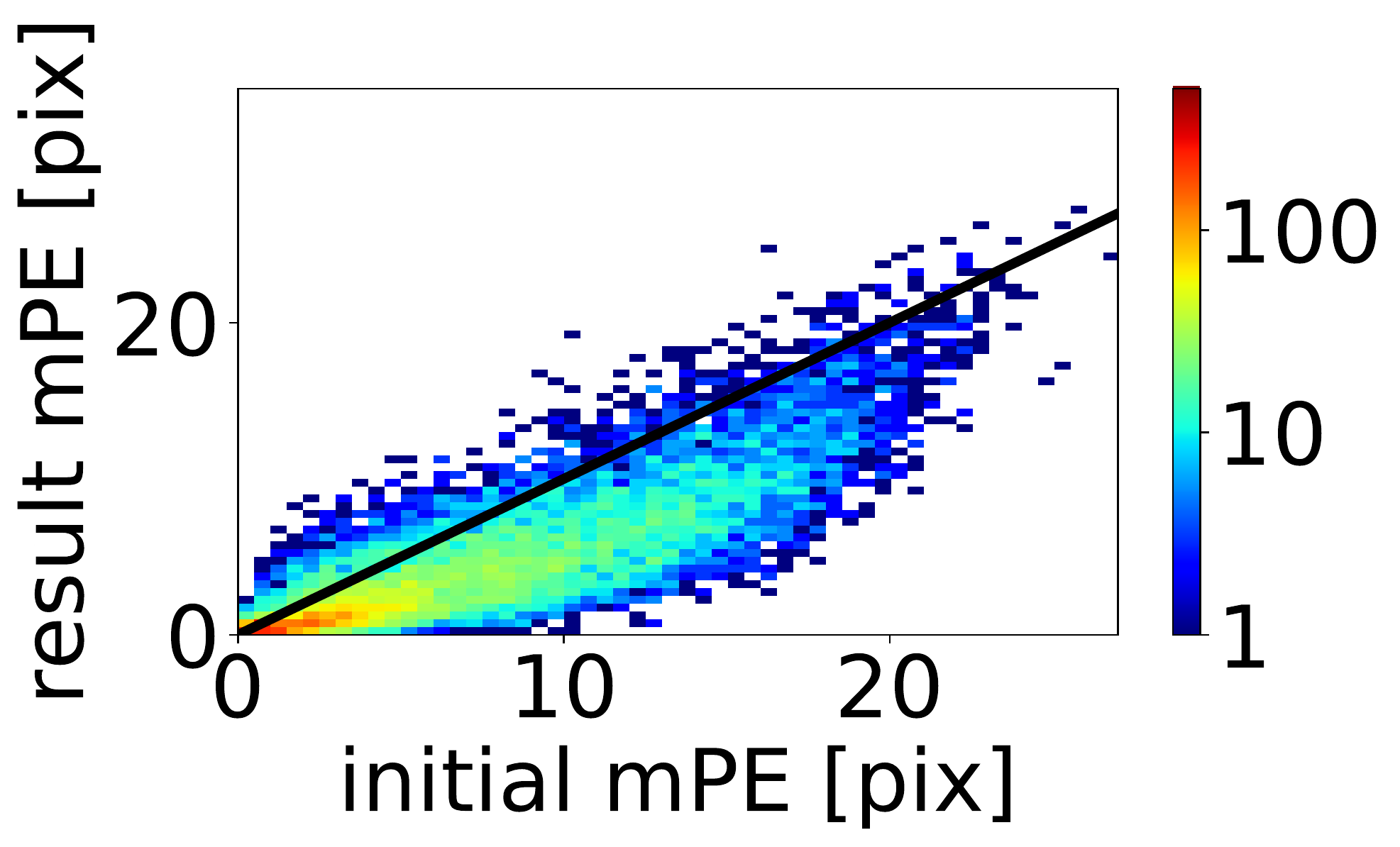}
     }
     \hfill
     \subfloat[\ppcl]{%
       \includegraphics[width=0.32\textwidth]{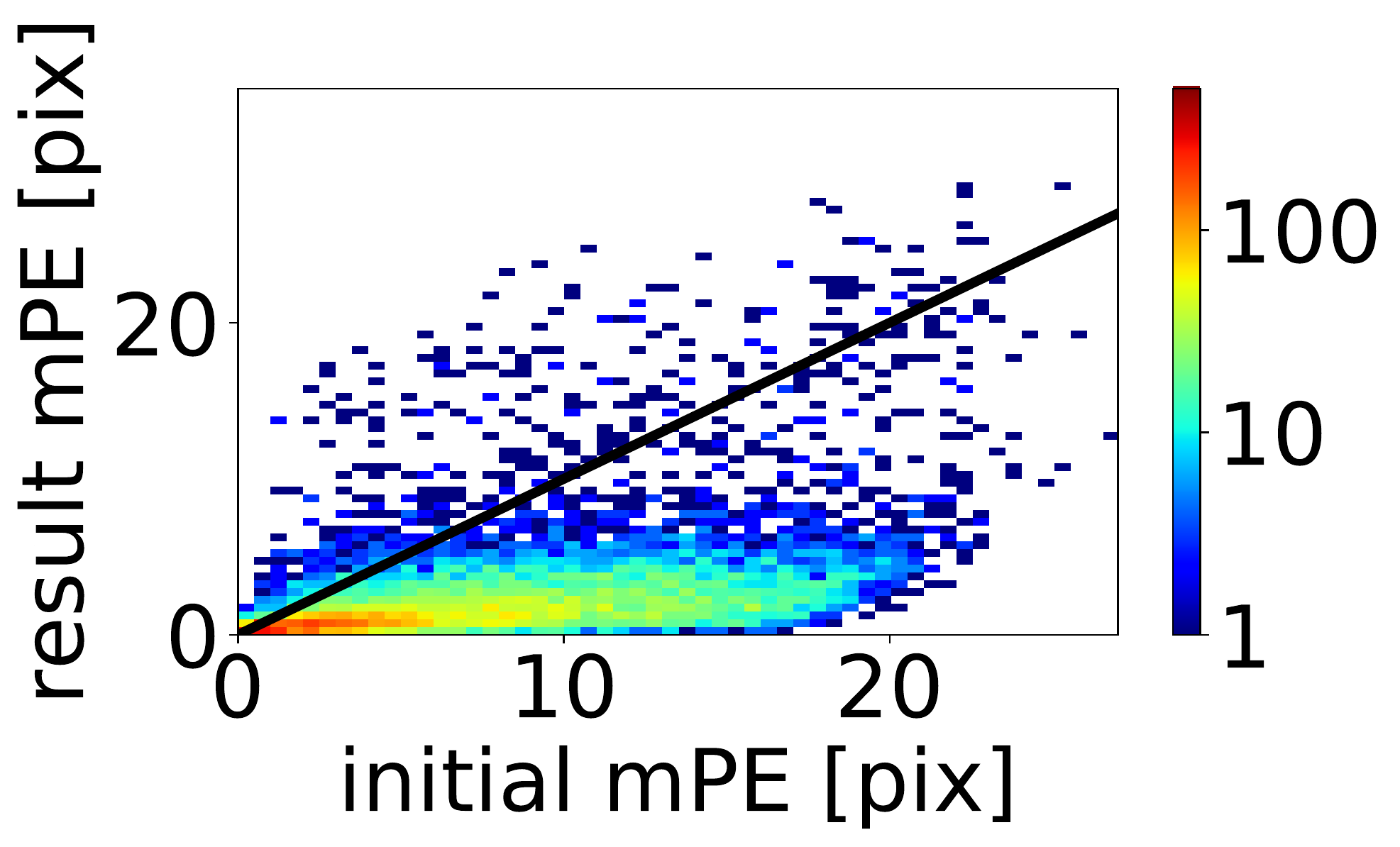}
     }
     \caption{Histograms showing initial and result projection error (PE) in pixels for a single iteration of registration on lowest resolution level (on validation set, 1024 correspondences per case). Motion estimation was performed using least squares for all methods. For PPC, no motion in depth is estimated (see Sec.~\ref{sec:comparedMethods}).}
     \label{fig:singleIter}
   \end{figure}
To better understand the effect of the correspondence weighting and regularization, we investigate the registration results after one iteration on the lowest resolution level. In Fig. \ref{fig:singleIter}, the PE in pixels (computed using $\set{\pointTarget_j}$ as target points) is shown for all cases in the validation set. As in training, 1024 correspondences are used per case for all methods. We observe that for PPC, the error has a high spread, where for some cases, it is decreased considerably, while for other cases, it is increased. For \ppcr, most cases are below the initial error. However, the error is decreased only marginally, as the regularization prevents large motions. For \ppcl, we observe that the error is drastically decreased for most cases. This shows that \ppcl is able to estimate motion efficiently. An example for correspondence weighting in \ppcl is shown in Fig.~\ref{fig:sample:W}, where we observe a set of consistent correspondences with high weights, while the remaining correspondences have low weights.

\subsubsection{Method Combinations}
\begin{figure}[t]
\centering
	 \subfloat[\ppcrmp]{%
       \includegraphics[width=0.49\textwidth]{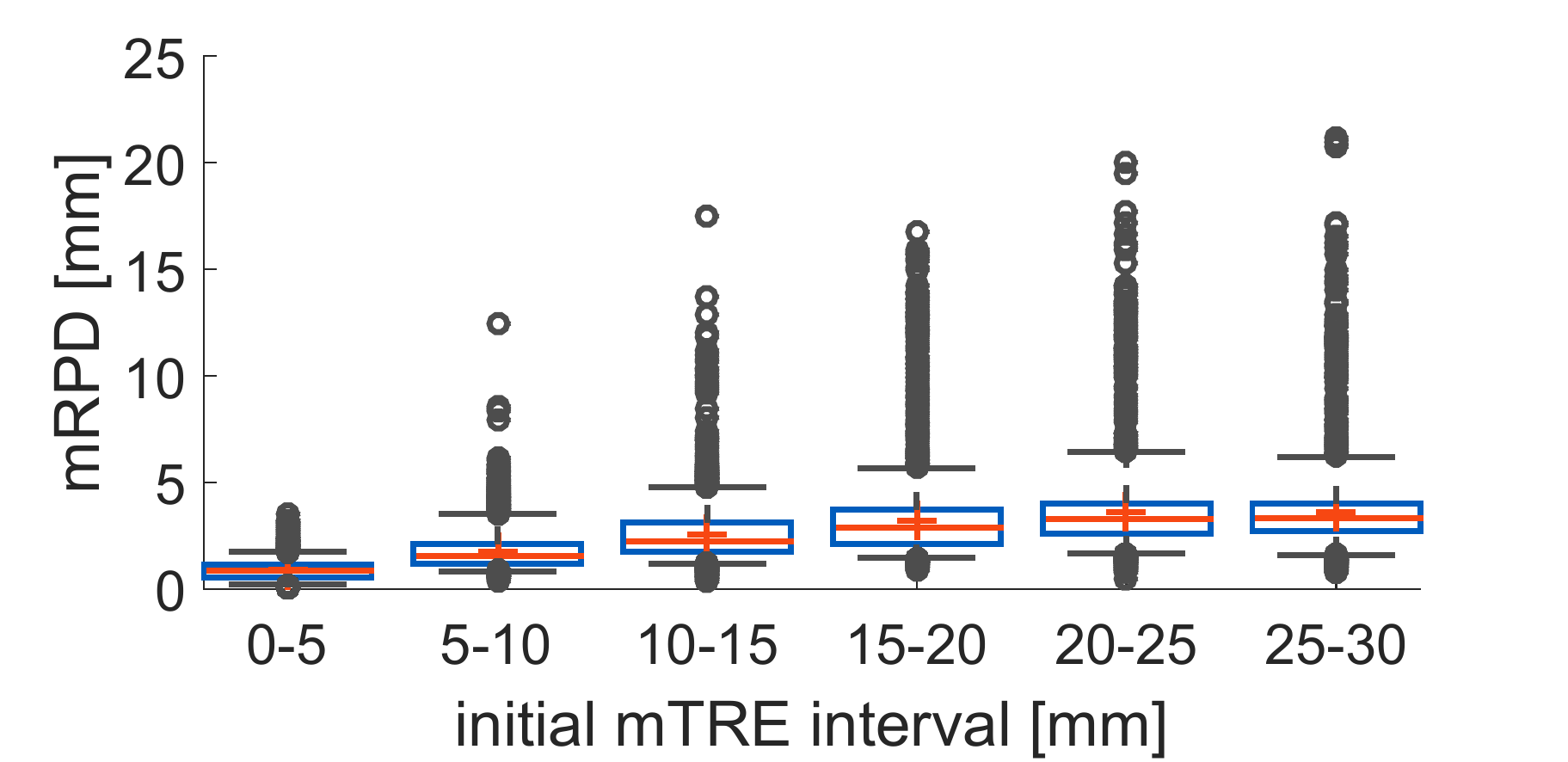}
     }
     \hfill
     \subfloat[\ppclp]{%
       \includegraphics[width=0.49\textwidth]{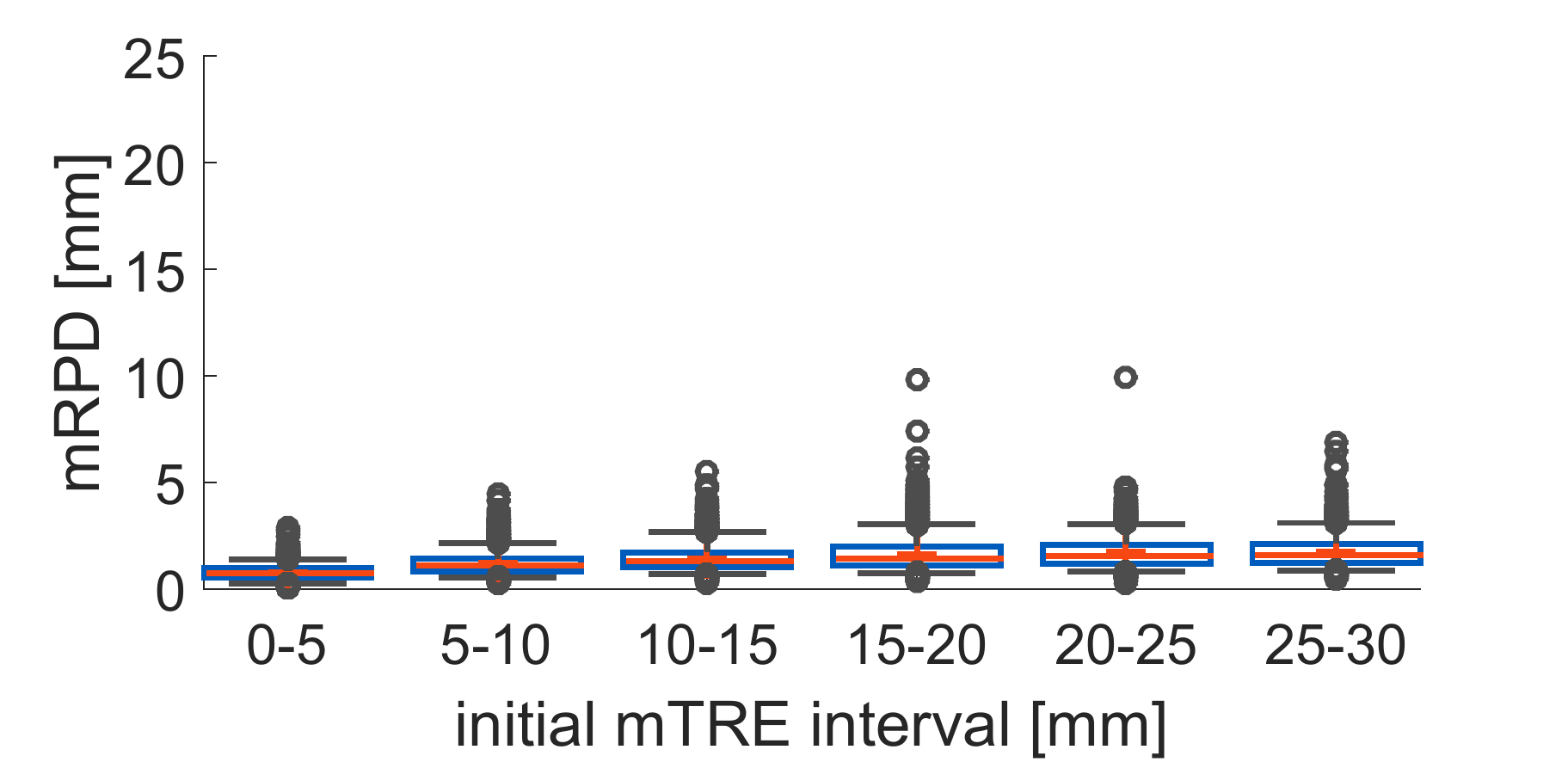}
     }
     \caption{Box plots for distribution of resulting mRPD on the lowest resolution level for successful registrations for different initial mTRE intervalls.}
     \label{fig:boxFirstResLevel}
\end{figure}
We observed that while the \ppcrm method has a high robustness (GCR and GSR), it leads to low accuracy. For \ppcl, we observed an increased GCR compared to the CR. In both cases, this demonstrates that registrations are present with a mRPD between 2\,mm and 10\,mm. As the PPC works reliably for small initial errors, we combine these methods with PPC by performing PPC on the highest resolution level instead of the respective method. We denote the resulting methods as \ppcrmp and \ppclp. We observe that \ppcrmp achieves an accuracy of 0.74$\pm$0.18\,mm, an SR of 94.6\,\% and a CR of 18\,mm, while \ppclp achieves an accuracy of 0.74$\pm$0.19\,mm, an SR of 96.1\,\% and a CR of 19\,mm. While the results are similar, we note that for \ppcrmp a manual weight selection is necessary. Further investigations are needed to clarify the better performance of PPC compared to \ppcl on the highest resolution level. However, this result may also demonstrate the strength of MCCR for cases where the majority of correspondences are correct.
We evaluate the convergence behavior of \ppclp and \ppcrmp by only considering cases which were successful. For these cases, we investigate the error distribution after the first resolution level. The results are shown in Fig. \ref{fig:boxFirstResLevel}. We observe that for \ppclp, a mRPD of below 10\,mm is achieved for all cases, while for \ppcrmp, higher misalignment of around 20\,mm mRPD is present. The result for \ppclp is achieved after an average of 7.6 iterations, while 11.8 iterations were performed on average for \ppcrmp using the stop criterion defined in~\cite{DRR17}. In combination, this further substantiates our findings from the single iteration evaluation and shows the efficiency of \ppcl and its potential for reducing the computational cost.

\section{Conclusion}
For \twothreed registration, we propose a method to learn the weighting of the local correspondences directly from the global criterion to minimize the registration error. We achieve this by incorporating the motion estimation and error computation steps into our training objective function. A modified PointNet network is trained to weight correspondences based on their geometrical properties and image similarity.
A large improvement in the registration robustness is demonstrated when using the \mbox{learning-based} correspondence weighting,  
while maintaining the high accuracy. Although a high robustness can also be achieved by regularized motion estimation, registration using learned correspondence weighting has the following advantages: it is more efficient, does not need manual parameter tuning and achieves a high accuracy.
One direction of future work is to further improve the weighting strategy, \eg~by including more information into the decision process and optimizing the objective function for robustness and/or accuracy depending on the stage of the registration, such as the current resolution level.
By regarding the motion estimation as part of the network and not the objective function, our model can also be understood in the framework of precision learning~\cite{PRT17} as a regression model for the motion, where we learn only the unknown component (weighting of correspondences), while employing prior knowledge to the known component (motion estimation). 
Following the framework of precision learning, replacing further steps of the registration framework with learned counterparts can be investigated. One candidate is the correspondence estimation, as it is challenging to design an optimal correspondence estimation method by hand.

{\bf Disclaimer:} The concept and software presented in this paper are based on research and are not commercially available. Due to regulatory reasons its future availability cannot be guaranteed.

\bibliographystyle{splncs04}
\bibliography{paper}

\end{document}